# Modeling Latent Variable Uncertainty for Loss-based Learning

**M. Pawan Kumar**  PAWAN.KUMAR@ECP.FR
Center for Visual Computing, École Centrale Paris, France
Équipe Galen, INRIA Saclay, Île-de-France, France
Université Paris-Est, LIGM (UMR CNRS), École des Ponts ParisTech, France

**Ben Packer**  BPACKER@CS.STANFORD.EDU
Stanford University, Stanford, CA 94305, USA

**Daphne Koller**  KOLLER@CS.STANFORD.EDU
Stanford University, Stanford, CA 94305, USA

## Abstract

We consider the problem of parameter estimation using weakly supervised datasets, where a training sample consists of the input and a partially specified annotation, which we refer to as the output. The missing information in the annotation is modeled using latent variables. Previous methods overburden a single distribution with two separate tasks: (i) modeling the uncertainty in the latent variables during training; and (ii) making accurate predictions for the output and the latent variables during testing. We propose a novel framework that separates the demands of the two tasks using two distributions: (i) a conditional distribution to model the uncertainty of the latent variables for a given input-output pair; and (ii) a delta distribution to predict the output and the latent variables for a given input. During learning, we encourage *agreement* between the two distributions by minimizing a loss-based dissimilarity coefficient. Our approach generalizes latent SVM in two important ways: (i) it models the uncertainty over latent variables instead of relying on a pointwise estimate; and (ii) it allows the use of loss functions that depend on latent variables, which greatly increases its applicability. We demonstrate the efficacy of our approach on two challenging problems—object detection and action detection—using publicly available datasets.

## 1. Introduction

Latent variable models (LVMs) provide an elegant formulation for learning with weakly supervised datasets.

For example, in computer vision, we may wish to learn a model for detecting an object such as 'deer' from images where the location of the deer is unknown, and is therefore treated as a latent variable. In computational medicine, we may wish to diagnose a patient based on the observed symptoms as well as other unknown factors, such as the family's medical history, which can be represented using latent variables.

Typically, an LVM employs a single distribution over three types of variables: (i) the observed variables, or input, whose values are known during both training and testing; (ii) the unobserved variables, or output, whose values are known only during training; and (iii) the unknown latent variables. In this setting, a natural framework would be to model the uncertainty in the value of the latent variables and learn an LVM by marginalizing them out (for example, in the case of the expectation-maximization, or EM, algorithm). However, such an approach is unsuited for applications that require an accurate prediction of the latent variables during test time. For example, in the above deer detection application, we would like to infer not only whether an image contains a deer, but also the exact location of the deer. Alternately, we can use a delta distribution that provides a pointwise estimate of the output and the latent variables (for example, in the case of the latent support vector machines, or LSVM, framework). However, discarding the uncertainty in latent variables can make such an approach prone to error due to noise (for example, background clutter that can be confused with a deer in feature space).

The above argument illustrates the deficiency of using a single joint distribution over the output and the latent variables to address two separate tasks: (i) modeling the uncertainty over latent variables during training; and (ii) making accurate predictions during testing. We address this deficiency by proposing a novel framework that consists of two distributions: (i) a conditional distribution to model the uncertainty of the latent variables for a given input-output pair; and (ii) a delta distribution to predict the output and the latent variables for a given input. In order to learn the distributions from a training dataset, we build on the intuition that they should *agree* with each other, that is, (i) the output predicted by the delta distribution should match the ground-truth output; and (ii) the latent variables predicted by the delta distribution should have a high probability according to the conditional distribution. Due to the limited representational power of any model we may not be able to achieve complete agreement (that is, all outputs are predicted correctly, and all predicted latent variables have probability 1). In order to make the two distributions as similar as possible, we minimize a regularized upper bound on a loss-based dissimilarity measure (Rao, 1982) between the distributions.

Unlike previous loss-based learning frameworks for LVMs, such as LSVM, we consider a general loss function that not only depends on the output but also the latent variables. Such a loss function is essential when solving problems that require the accurate prediction of latent variables (for example, the aforementioned object detection problem). By not restricting the form of the loss function, our framework greatly enhances the applicability of loss-based learning with latent variables. In fact, our framework can be viewed as a strict generalization of LSVM in the sense that, when the loss function is independent of the true (unknown) value of the latent variables, it reduces to an LSVM.

Throughout this paper, we will assume that the latent variables are helpful in predicting the correct output of a sample. For example, if we want to distinguish between images of deers and elephants, we would expect that the background clutter to have similar appearance for both categories, and an accurate object localization to be essential for correct prediction. There may be cases where this assumption does not hold. For example, images of deers and cars could be distinguished by detecting roads, or other objects that are more commonly found in urban environments. However, even in such cases, we may be able to learn to detect the object by providing fully supervised annotations for a small fraction of training images, which would help guide the learner towards the correct object locations in other weakly supervised training images.

## 2. Related Work

The most commonly used method for learning the parameters of an LVM is the EM algorithm (Dempster et al., 1977; Sundberg, 1974), or its many variants (Gelman et al., 1995). While the EM algorithm has an elegant probabilistic interpretation of maximizing the likelihood of the ground-truth output, it marginalizes out the latent variables, which makes it unsuited to problems that require the accurate prediction of latent variables. Furthermore, it does not employ a user-specified loss function, which captures the user's assessment of the quality of the solution.

The most related works to our approach are LSVM (Felzenszwalb et al., 2008; Yu & Joachims, 2009) and its recently proposed generalization called max-margin min-entropy models (or M3E for short) (Miller et al., 2012). The parameters of an LSVM or an M3E are learned by minimizing a regularized upper bound of the training loss. However, the loss function is restricted to be independent of the true (unknown) value of the latent variables. While such loss functions are useful, and in fact have been successfully employed in practice (Blaschko et al., 2010; Felzenszwalb et al., 2008; Kumar et al., 2010; Miller et al., 2012; Yu & Joachims, 2009), they cannot model several important problems, including the two employed in our experiments—object detection and action detection. In contrast, our framework allows the use of a general loss function. In section 4 we will show that, for loss functions that are independent of the true value of the latent variable, our framework reduces to an LSVM.

In our earlier work (Kumar et al., 2011), we proposed an iterative LSVM strategy (or ILSVM for short) with the aim of using a general loss function. In section 5, we show that ILSVM corresponds to using delta functions to model the conditional distribution of the latent variables given the input and the output. In our experiments, we show that using a non-delta conditional distribution significantly outperforms ILSVM.

## 3. Preliminaries

**Notation.** We denote the input by $\mathbf{x} \in \mathcal{X}$, the output by $\mathbf{y} \in \mathcal{Y}$ and the latent variables by $\mathbf{h} \in \mathcal{H}$. The training dataset $\mathcal{D} = \{\mathbf{s}_i = (\mathbf{x}_i, \mathbf{y}_i), i = 1, \cdots, n\}$ consists of $n$ input-output pairs (or samples) $\mathbf{s}_i$.

We denote the parameters of the delta distribution, which predicts the output and the latent variables for a given input, as $\mathbf{w}$. The parameters of the conditional distribution of the latent variables given the input and the output are denoted by $\boldsymbol{\theta}$.

We assume that the user specifies a loss function $\Delta(\mathbf{y}_1, \mathbf{h}_1, \mathbf{y}_2, \mathbf{h}_2)$ that measures the difference between $(\mathbf{y}_1, \mathbf{h}_1)$ and $(\mathbf{y}_2, \mathbf{h}_2)$. Similar to previous approaches, we assume that $\Delta(\mathbf{y}_1, \mathbf{h}_1, \mathbf{y}_2, \mathbf{h}_2) = 0$ if $\mathbf{y}_1 = \mathbf{y}_2$ and $\mathbf{h}_1 = \mathbf{h}_2$. Otherwise, $\Delta(\mathbf{y}_1, \mathbf{h}_1, \mathbf{y}_2, \mathbf{h}_2) \geq 0$.

**Rao's Dissimilarity Coefficient.** We provide a brief description of the dissimilarity measure used in our framework, which was first introduced by Rao (1982). Given a loss function $\Delta(\mathbf{z}_1, \mathbf{z}_2)$, where $\mathbf{z}_1, \mathbf{z}_2 \in \mathcal{Z}$, the diversity coefficient of two distributions $\mathrm{P}_i(\mathbf{z})$ and $\mathrm{P}_j(\mathbf{z})$ is defined as the expected loss between two samples drawn randomly from the two distributions respectively, that is,

$$H(\mathrm{P}_i, \mathrm{P}_j) = \sum_{\mathbf{z}_1 \in \mathcal{Z}} \sum_{\mathbf{z}_2 \in \mathcal{Z}} \Delta(\mathbf{z}_1, \mathbf{z}_2) \, \mathrm{P}_i(\mathbf{z}_1) \, \mathrm{P}_j(\mathbf{z}_2). \quad (1)$$

Using the diversity coefficient, the dissimilarity coefficient between the two distributions can be defined as the following Jensen difference:

$$D(\mathrm{P}_i, \mathrm{P}_j) = H(\mathrm{P}_i, \mathrm{P}_j) - \beta H(\mathrm{P}_i, \mathrm{P}_i) - (1-\beta) H(\mathrm{P}_j, \mathrm{P}_j), \quad (2)$$

where $\beta \in (0, 1)$. Note that Rao (1982) fixed $\beta = 0.5$ in order to ensure that the dissimilarity coefficient is symmetric for $\mathrm{P}_i$ and $\mathrm{P}_j$. However, dissimilarity coefficients do not necessarily have to be symmetric (for example, the well-known Kullback-Liebler divergence is non-symmetric); hence we use the more general version shown in equation (2). Rao (1982) showed that the above formulation generalizes other commonly used dissimilarity coefficients such as the Mahalanobis distance and the Gini-Simpson index. We refer the reader to (Rao, 1982) for details.

## 4. Loss-based Learning Framework

Using the above notation and definitions, we now provide the details of our learning framework. We begin by describing the distributions represented by the LVM.

### 4.1. Distributions

We wish to address two separate tasks: (i) to accurately model the distribution of the latent variables for a given input-output pair; and (ii) to accurately predict the output and latent variables for a given input (where accuracy is measured by a user-defined loss). Instead of addressing these two tasks with a single distribution as in previous works, we define two separate distributions, each focused on a single task.

Given an input $\mathbf{x}$, we define a delta distribution parameterized by $\mathbf{w}$ that predicts the output and the latent variables according to the following rule:

$$(\mathbf{y}(\mathbf{w}), \mathbf{h}(\mathbf{w})) = \operatorname*{argmax}_{(\mathbf{y}, \mathbf{h})} \mathbf{w}^\top \Psi(\mathbf{x}, \mathbf{y}, \mathbf{h}). \quad (3)$$

Here, $\Psi(\mathbf{x}, \mathbf{y}, \mathbf{h})$ is a joint feature vector of the input $\mathbf{x}$, the output $\mathbf{y}$ and the latent variables $\mathbf{h}$. Note that, although for simplicity we defined a linear rule in $\mathbf{w}$, we can also employ a non-linear kernel within our framework. Formally, the delta distribution is given by

$$\mathrm{P}_\mathbf{w}(\mathbf{y}, \mathbf{h}|\mathbf{x}) = \begin{cases} 1 & \text{if } \mathbf{y} = \mathbf{y}(\mathbf{w}), \mathbf{h} = \mathbf{h}(\mathbf{w}), \\ 0 & \text{otherwise.} \end{cases} \quad (4)$$

As mentioned earlier, since the true value of the latent variables is unknown, we would like to model the uncertainty in their values. To this end, we define a separate conditional distribution parameterized by $\boldsymbol{\theta}$ such that

$$\mathrm{P}_{\boldsymbol{\theta}}(\mathbf{h}_i|\mathbf{s}_i) = \frac{1}{Z(\mathbf{s}_i; \boldsymbol{\theta})} \exp\left(\boldsymbol{\theta}^\top \Phi(\mathbf{x}_i, \mathbf{y}_i, \mathbf{h}_i)\right), \quad (5)$$

where $Z(\mathbf{s}_i; \boldsymbol{\theta})$ is the partition function, which ensures that the distribution sums to one and $\Phi(\mathbf{x}_i, \mathbf{y}_i, \mathbf{h}_i)$ is a joint feature vector of the input $\mathbf{x}_i$, the output $\mathbf{y}_i$ and the latent variables $\mathbf{h}_i$. This feature vector can be different than the joint feature vector used to specify the delta distribution $\mathrm{P}_\mathbf{w}(\cdot)$. Once again, a log-linear distribution is used only to simplify the description. Our framework is valid for any general form of the distribution $\mathrm{P}_{\boldsymbol{\theta}}(\cdot)$. Using the above conditional distribution, we also specify a joint distribution as follows:

$$\mathrm{P}'_{\boldsymbol{\theta}}(\mathbf{y}, \mathbf{h}_i|\mathbf{x}_i) = \begin{cases} \mathrm{P}_{\boldsymbol{\theta}}(\mathbf{h}_i|\mathbf{s}_i) & \text{if } \mathbf{y} = \mathbf{y}_i, \\ 0 & \text{otherwise.} \end{cases} \quad (6)$$

As will be seen shortly, this joint distribution would allow us to employ Rao's dissimilarity coefficient in our learning framework.

### 4.2. The Learning Objective

Given a dataset $\mathcal{D}$ and a loss function $\Delta(\cdot)$, we propose to learn the parameters $\mathbf{w}$ and $\boldsymbol{\theta}$ such that it minimizes the corresponding dissimilarity coefficient over all training samples. Before delving into the details, we give a broad overview of our objective function.

For a fixed $\mathbf{w}$, if the predicted output $\mathbf{y}_i(\mathbf{w})$ is similar to the ground-truth output $\mathbf{y}_i$, our objective encourages the probability of the corresponding latent variables, that is $\mathrm{P}_{\boldsymbol{\theta}}(\mathbf{h}_i(\mathbf{w})|\mathbf{s}_i)$, and other similar latent variables, to be high. If the predicted output $\mathbf{y}_i(\mathbf{w})$ is dissimilar to the ground-truth output $\mathbf{y}_i$, our objective encourages the diversity coefficient of the corresponding distribution, that is $\mathrm{P}_{\boldsymbol{\theta}}(\cdot|\mathbf{s}_i)$, to be high. In other words, for a correctly predicting sample, the conditional distribution is *peaky*, while for an incorrectly predicted sample, the conditional distribution is *flat*.

For a fixed $\boldsymbol{\theta}$, our objective minimizes the *expected loss* of the prediction $(\mathbf{y}_i(\mathbf{w}), \mathbf{h}_i(\mathbf{w}))$ over all the training samples $\mathbf{s}_i$. This is a key point of our formulation, as the expected loss incorporates the uncertainty of the latent variable values while learning the parameters $\mathbf{w}$. Formally, the expected loss of a pair of output and latent variables $(\mathbf{y}, \mathbf{h})$ for the sample $\mathbf{s}_i$, measured with respect to $P_{\boldsymbol{\theta}}(\cdot|\mathbf{s}_i)$, is defined as

$$\Delta_i(\mathbf{y}, \mathbf{h}; \boldsymbol{\theta}) = \sum_{\mathbf{h}_i} \Delta(\mathbf{y}_i, \mathbf{h}_i, \mathbf{y}, \mathbf{h}) P_{\boldsymbol{\theta}}(\mathbf{h}_i|\mathbf{s}_i), \quad (7)$$

that is, it is the expectation of the loss between $(\mathbf{y}, \mathbf{h})$ and $(\mathbf{y}_i, \mathbf{h}_i)$, where the expectation is taken over the distribution of the unknown latent variables $\mathbf{h}_i$.

We now provide a mathematical description of our learning framework. However, throughout this section and the next, we will reiterate the above intuition at the appropriate places. Our training objective is the sum of the dissimilarity coefficient between $P_\mathbf{w}(\cdot)$ and $P'_{\boldsymbol{\theta}}(\cdot)$ over all training samples. Using the definition of dissimilarity coefficient in equation (2), the objective can be written in terms of expected loss as

$$D(\mathbf{w}, \boldsymbol{\theta}) = \frac{1}{n}\left(\sum_{i=1}^{n} H_i(\mathbf{w}, \boldsymbol{\theta}) - \beta H_i(\boldsymbol{\theta})\right),$$
$$H_i(\mathbf{w}, \boldsymbol{\theta}) = \Delta_i(\mathbf{y}_i(\mathbf{w}), \mathbf{h}_i(\mathbf{w}); \boldsymbol{\theta}),$$
$$H_i(\boldsymbol{\theta}) = \sum_{\mathbf{h}'_i} P_{\boldsymbol{\theta}}(\mathbf{h}'_i|\mathbf{s}_i)\Delta_i(\mathbf{y}_i, \mathbf{h}'_i; \boldsymbol{\theta}). \quad (8)$$

Note that the diversity coefficient of $P_\mathbf{w}(\cdot)$ is 0 since it is a delta distribution. Hence, the term $H_i(\mathbf{w})$ vanishes from the above objective.

Minimizing the objective (8) encourages two desirable properties: (i) the predicted output $\mathbf{y}_i(\mathbf{w})$ should be similar to the ground-truth output $\mathbf{y}_i$; and (ii) the predicted latent variable $\mathbf{h}_i(\mathbf{w})$ should be similar to the latent variables with high probabilities $P_{\boldsymbol{\theta}}(\mathbf{h}_i|\mathbf{x}_i, \mathbf{y}_i)$. Importantly, the similarity (or, to be more precise, the dissimilarity) of the outputs and the latent variables is specified by the loss function $\Delta(\cdot)$. Hence, during learning, the parameters $\mathbf{w}$ and $\boldsymbol{\theta}$ are tuned according to the user's domain knowledge regarding the quality of a solution. This ability to learn loss-specific parameters is absent in traditional frameworks such as EM and its variants.

### 4.3. Upper Bound on the Learning Objective

While the objective (8) is smooth and differentiable in $\boldsymbol{\theta}$, for most commonly used choices of the loss function it is highly non-smooth in $\mathbf{w}$. The non-smoothness of the objective results in a difficult optimization problem, which makes the learner prone to bad local minimum solutions. In order to overcome this deficiency, we minimize an upper bound on the objective, similar to the LSVM formulation (Yu & Joachims, 2009).

Specifically, we upper bound the term $H_i(\mathbf{w}, \boldsymbol{\theta})$, which depends on $\mathbf{w}$, using $\xi_i(\mathbf{w}, \boldsymbol{\theta})$ defined as follows.

$$\xi_i(\mathbf{w}, \boldsymbol{\theta}) \leq \max_{\mathbf{y}, \mathbf{h}} \left\{ \mathbf{w}^\top \Psi(\mathbf{x}_i, \mathbf{y}, \mathbf{h}) + \Delta_i(\mathbf{y}, \mathbf{h}; \boldsymbol{\theta}) \right\} \\ - \max_{\mathbf{h}} \mathbf{w}^\top \Psi(\mathbf{x}_i, \mathbf{y}_i, \mathbf{h}) \quad (9)$$

Using the above inequalities, the objective $D(\mathbf{w}, \boldsymbol{\theta})$ can be upper bounded as

$$U(\mathbf{w}, \boldsymbol{\theta}) = \frac{1}{n}\left(\sum_{i=1}^{n}\xi_i(\mathbf{w}, \boldsymbol{\theta}) - \beta H_i(\boldsymbol{\theta})\right). \quad (10)$$

However, if we learn the parameters $\mathbf{w}$ and $\boldsymbol{\theta}$ by minimizing the above upper bound (or indeed the original objective function), we run the risk of overfitting to the training data. In order to prevent this, we introduce regularization terms for the parameters. For this work, we use $\ell_2$ norms, though other norms may also be employed. To summarize, the parameters are learned by solving the following optimization problem:

$$(\mathbf{w}^*, \boldsymbol{\theta}^*) = \underset{(\mathbf{w}, \boldsymbol{\theta})}{\operatorname{argmin}} \frac{1}{2}||\mathbf{w}||^2 + \frac{J}{2}||\boldsymbol{\theta}||^2 + CU(\mathbf{w}, \boldsymbol{\theta}), \quad (11)$$

where the hyperparameters $J$ and $C$ are the relative weights for the regularization of $\boldsymbol{\theta}$ and the upper bound of the dissimilarity coefficient respectively. Note that the upper bound derivation and the resulting optimization problem are similar to the LSVM framework. In fact, the problem can be shown to be a strict generalization of LSVM.

**Observation 1.** When the loss function does not depend on the value of the latent variables, problem (11) is equivalent to the problem of learning an LSVM.

This observation follows from the fact that, when the loss function is independent of the latent variables, $H_i(\boldsymbol{\theta}) = \Delta_i(\mathbf{y}_i; \boldsymbol{\theta}) \sum_{\mathbf{h}'_i} P_{\boldsymbol{\theta}}(\mathbf{h}'_i|\mathbf{s}_i) = 0$. Hence, the optimization problem is equivalent to minimizing the sum of the regularization of $\mathbf{w}$ and $\xi_i(\mathbf{w}, \boldsymbol{\theta})$ (which are equivalent to the slack variables that model the upper bound of the loss function for the sample $\mathbf{s}_i$ in LSVM). In fact, even if the loss function does depend on the predicted latent variable $\mathbf{h}_i(\mathbf{w})$, the optimization problem (11) still generalizes LSVM. This follows from the fact that, in this case, the LSVM problem is equivalent to using delta distributions to model $P_{\boldsymbol{\theta}}(\cdot)$. Formal proofs are omitted.

## 5. Optimization

While the upper bound derived in the previous section still results in a non-smooth and non-convex optimization problem, we obtain an approximate solution using block coordinate descent. Specifically, starting with some initial estimate of parameters, we alternately fix one of the two sets of parameters (either $\mathbf{w}$ or $\boldsymbol{\theta}$) while optimizing problem (11) over the other set of parameters. The process is said to terminate when the decrease in the objective falls below $C\epsilon$, where $C$ is the hyperparameter in problem (11) and $\epsilon$ is a user specified tolerance. The following subsections provide the details of the optimization over each set of parameters.

### 5.1. Optimization over $\mathbf{w}$

For a fixed $\boldsymbol{\theta}$, problem (11) can be interpreted as minimizing a regularized upper bound on the expected loss induced by $\mathbf{w}$, that is,

$$\sum_i \Delta_i(\mathbf{y}_i(\mathbf{w}), \mathbf{h}_i(\mathbf{w}); \boldsymbol{\theta}), \qquad (12)$$

since the term $H_i(\boldsymbol{\theta})$ is a constant for all samples $\mathbf{s}_i$. The expected loss is an intuitive objective: it gives more weight to the loss corresponding to the latent variables that have a high probability and less weight to those corresponding to the latent variables with low probability. Formally, for a fixed $\boldsymbol{\theta}$, the optimization problem (11) reduces to the following:

$$\begin{aligned}\min_{\mathbf{w}} \quad & \tfrac{1}{2}||\mathbf{w}||^2 + \tfrac{C}{n}\sum_i \xi_i \\ \text{s.t.} \quad & \xi_i = \max_{\mathbf{y},\mathbf{h}}\left\{\mathbf{w}^\top \Psi(\mathbf{x}_i, \mathbf{y}, \mathbf{h}) + \Delta_i(\mathbf{y}, \mathbf{h}; \boldsymbol{\theta})\right\} \\ & \quad - \max_{\mathbf{h}} \mathbf{w}^\top \Psi(\mathbf{x}_i, \mathbf{y}_i, \mathbf{h}). \end{aligned} \qquad (13)$$

The following observation provides us with an efficient algorithm for the above optimization problem.

**Observation 2.** Problem (13) is a difference-of-convex program.

The regularization term $||\mathbf{w}||^2$ is convex. The term $\xi_i(\mathbf{w}, \boldsymbol{\theta})$ is a difference of two functions that are the pointwise maximum of a set of linear functions. Since the pointwise maximum of convex functions is convex, the observation follows. Similar to LSVM, a local minimum or saddle point solution of problem (13) can be found using the concave-convex procedure (CCCP) (Yu & Joachims, 2009). The main steps of CCCP are outlined in Algorithm 1. It iteratively estimates the value of the latent variables using the current estimate of $\mathbf{w}$, and updates the parameters by solving a convex optimization problem (14). There are several efficient algorithms for problem (14), for example (Joachims et al., 2009; Shalev-Shwartz et al., 2009; Tsochantaridis et al., 2004). In this work, we use the 1-slack reformulation method proposed by Joachims et al. (2009). We can also solve problem (13) using the self-paced learning algorithm (Kumar et al., 2010), which can potentially improve the performance of our framework. However, in this paper, we restrict ourselves to the simpler and more efficient CCCP algorithm.

---

**Algorithm 1** The CCCP algorithm for optimizing $\mathbf{w}$.
**input** Dataset $\mathcal{D}$, initial estimate $\mathbf{w}_0$, tolerance $\epsilon$.
1: $t \leftarrow 0$.
2: **repeat**
3:    Update $\mathbf{h}_i^* = \text{argmax}_{\mathbf{h}_i} \mathbf{w}_t^\top \Psi(\mathbf{x}_i, \mathbf{y}_i, \mathbf{h}_i)$.
4:    Estimate the updated parameter $\mathbf{w}_{t+1}$ by solving the following convex optimization problem:

$$\begin{aligned}\min_{\mathbf{w}} \quad & \tfrac{1}{2}||\mathbf{w}||^2 + \tfrac{C}{n}\sum_i \xi_i \\ \text{s.t.} \quad & \xi_i \geq \mathbf{w}^\top \Psi(\mathbf{x}_i, \mathbf{y}, \mathbf{h}) + \Delta_i(\mathbf{y}, \mathbf{h}; \boldsymbol{\theta}) \\ & \quad - \mathbf{w}^\top \Psi(\mathbf{x}_i, \mathbf{y}_i, \mathbf{h}_i^*), \forall \mathbf{y}, \mathbf{h}. \end{aligned} \qquad (14)$$

5:    $t \leftarrow t+1$.
6: **until** Objective cannot be decreased below $C\epsilon$.

---

Problem (13) requires the computation of the expected loss $\Delta_i(\mathbf{y}, \mathbf{h}; \boldsymbol{\theta})$ as defined in equation (7), which can be found in $O(|\mathcal{H}|)$ time for each pair of $(\mathbf{y}, \mathbf{h})$ (where $\mathcal{H}$ is the space of all latent variables). For a sufficiently small $\mathcal{H}$ this operation is computationally feasible. For a large latent variable space $\mathcal{H}$, we have two options. First, we can choose the joint feature vector $\Phi(\mathbf{x}, \mathbf{y}, \mathbf{h})$ for the conditional distribution $P_{\boldsymbol{\theta}}(\cdot)$ to be decomposable in such a manner as to facilitate efficient computation of sufficient statistics (for example, a low tree-width model). Note that this still allows us to use a more complex joint feature vector $\Psi(\mathbf{x}, \mathbf{y}, \mathbf{h})$ to make predictions for a given test sample. Second, if the problem requires a complex $\Phi(\mathbf{x}, \mathbf{y}, \mathbf{h})$ to encode the conditional distribution, then we can resort to using one of several inference techniques to compute the approximate sufficient statistics. However, we note that several important problems in machine learning can be formulated using latent variables whose space is sufficiently small to allow for exact computations of the expected loss, including motif finding (Yu & Joachims, 2009), image classification (Kumar et al., 2010; Miller et al., 2012), digit recognition (Kumar et al., 2010), and the two problems used in our experiments, namely object detection and action detection.

### 5.2. Optimization over $\boldsymbol{\theta}$

For a fixed $\mathbf{w}$, problem (11) can be interpreted as a regularized upper bound on the following objective

$$\frac{1}{n}\left(\sum_{i=1}^n H_i(\mathbf{w}, \boldsymbol{\theta}) - \beta H_i(\boldsymbol{\theta})\right), \qquad (15)$$

where the divergence coefficients $H_i(\mathbf{w}, \boldsymbol{\theta})$ and $H_i(\boldsymbol{\theta})$ are defined in equation (8). To gain an understanding of the above objective, let us consider a simple 0/1 loss (that is, the loss is 0 if both the outputs are equal and both the latent variables are equal, otherwise 1). If $\mathbf{y}_i(\mathbf{w}) = \mathbf{y}_i$, that is, $\mathbf{w}$ predicts the correct output for the sample $\mathbf{s}_i$, then the first term of the above objective dominates the second. In this case, the parameter $\boldsymbol{\theta}$ is encouraged to assign a high probability to the predicted latent variables $\mathbf{h}_i(\mathbf{w})$, and other similar latent variables, in order to minimize the objective. If $\mathbf{y}_i(\mathbf{w}) \neq \mathbf{y}_i$, the first term is a constant. Thus, the parameter $\boldsymbol{\theta}$ is encouraged to maximize the diversity of the conditional distribution $P_{\boldsymbol{\theta}}(\cdot)$. In other words, for a correct prediction of output, we learn a *peaky* distribution and for an incorrect prediction of output, we learn a *flat* distribution. Formally, for a fixed $\mathbf{w}$, the optimization problem (11) reduces to the following:

$$\min_{\boldsymbol{\theta}} \frac{J}{2} ||\boldsymbol{\theta}||^2 + CU(\mathbf{w}, \boldsymbol{\theta}), \quad (16)$$

where $U(\mathbf{w}, \boldsymbol{\theta})$ is defined in equation (10). We obtain an approximate solution to the above problem using stochastic subgradient descent (SSD). The main steps of SSD are outlined in Algorithm 2.

**Algorithm 2** The SSD algorithm for optimizing $\boldsymbol{\theta}$.
**input** Dataset $\mathcal{D}$, initial estimate $\boldsymbol{\theta}_0$, $T > 0$.
1: $t \leftarrow 0$. $\lambda \leftarrow J/C$.
2: **repeat**
3:   Choose a sample $\mathbf{s}_i$ randomly from $\mathcal{D}$.
4:   Compute the stochastic subgradient $\mathbf{g}_t$ as

$$\mathbf{g}_t = \boldsymbol{\theta}_t + \nabla_{\boldsymbol{\theta}} H_i(\mathbf{w}, \boldsymbol{\theta}) + \nabla_{\boldsymbol{\theta}} H_i(\boldsymbol{\theta}). \quad (17)$$

5:   $t \leftarrow t + 1$.
6:   Update $\boldsymbol{\theta}_{t+1} \leftarrow \boldsymbol{\theta}_t - \frac{1}{\lambda t} \mathbf{g}_t$.
7: **until** Number of iterations $t = T$.

Each iteration of SSD takes $O(|\mathcal{H}|^2)$ time (since the subgradient $\mathbf{g}_t$ requires a quadratic sum to compute $H_i(\boldsymbol{\theta})$). Similar to the expected loss, this can be performed exactly for a sufficiently small space of latent variables, or the appropriate choice of the joint feature vector $\Phi(\mathbf{x}, \mathbf{y}, \mathbf{h})$. For a large latent variable space and a complex joint feature vector, we would have to resort to approximate inference.

### 5.3. Comparison with ILSVM

Our overall approach is similar in flavor to the ILSVM algorithm (Kumar et al., 2011), which iterates over the following two steps until convergence: (i) obtain the value of the latent variables for all training samples using the current estimate of the parameters; (ii) update the parameters by solving an LSVM, where the loss function is measured using the latent variables estimated in the first step instead of the true latent variables. The following observation shows that ILSVM is a special case of our framework.

**Observation 3.** The first step of ILSVM minimizes the objective (15) when $P_{\boldsymbol{\theta}}(\cdot)$ are restricted to be delta distributions. The second step of ILSVM solves an LSVM problem similar to the one described in the previous subsection for optimizing over $\mathbf{w}$.

The observation regarding the second step is straightforward. For the first step, it follows from the fact that ILSVM minimizes $H_i(\mathbf{w}, \boldsymbol{\theta})$. As the second divergence coefficient $H_i(\boldsymbol{\theta})$ vanishes when using delta conditional distributions, ILSVM effectively minimizes objective (15) for a fixed $\mathbf{w}$. A formal proof is omitted.

## 6. Experiments

We now demonstrate the efficacy of our framework on two challenging machine learning applications: object detection and action detection. Specifically, we show how our approach, which models the uncertainty in the values of the latent variables during training, outperforms the previous loss-based learning frameworks, namely LSVM and ILSVM, which only estimate the most likely assignment of the latent variables. All three methods used in our experiments share a common hyperparameter $C$ (the relative weight for the upper bounds $\xi_i$), which we vary to take values from the set $\{10^{-4}, 10^{-3}, \cdots, 10^2\}$. In addition, our framework introduces two more hyperparameters: $J$ (the relative weight for the regularization of $\boldsymbol{\theta}$) and $\beta$ (the hyperparameter for Rao's dissimilarity coefficient). In all our experiments, we set $J = 0.1$ and $\beta = 0.1$. However, we may obtain better results by carefully tuning these hyperparameters. The tolerance value for all the methods was set to $\epsilon = 10^{-3}$.

### 6.1. Object Detection

**Problem Formulation.** The aim of this application is to learn discriminative object models that predict the category (for example, 'deer' or 'elephant') and the location of the object present in an image. In a fully supervised setting, we would be required to specify a tight bounding box around the object present in each of the training samples. As the collection of such annotations is onerous and expensive, we would like to learn the object models using image-level labels (that is, labels indicating the presence or absence of an object category in an image), which are considerably easier to obtain. Formally, for each sample, the input $\mathbf{x}$ is an image. The output $\mathbf{y} \in \{0, 1, \cdots, c-1\}$, where $c$ is the number of object categories. The latent variable $\mathbf{h}$ models the tight bounding box around the object in the image. Similar to previous works (Kumar et al., 2010; Miller et al., 2012), the joint feature vectors $\Psi(\mathbf{x}, \mathbf{y}, \mathbf{h})$ and $\Phi(\mathbf{x}, \mathbf{y}, \mathbf{h})$ are defined using the HOG descriptor (Dalal & Triggs, 2005; Felzenszwalb et al., 2008) extracted using the pixels of the bounding box. In our experiments, we consider non-overlapping putative bounding boxes that are 8 pixels apart, which results in a maximum of 350 bounding boxes for each image in our dataset. This allows us to compute the exact expected loss and the exact subgradients during learning. We employ two different loss functions, 0/1 loss and overlap loss, which are defined below.

$$\Delta_{0/1}(\mathbf{y}_1, \mathbf{h}_1, \mathbf{y}_2, \mathbf{h}_2) = \begin{cases} 0 & \text{if } \mathbf{y}_1 = \mathbf{y}_2, \mathbf{h}_1 = \mathbf{h}_2, \\ 1 & \text{otherwise,} \end{cases}$$

$$\Delta_O(\mathbf{y}_1, \mathbf{h}_1, \mathbf{y}_2, \mathbf{h}_2) = \begin{cases} 1 - O(\mathbf{h}_1, \mathbf{h}_2) & \text{if } \mathbf{y}_1 = \mathbf{y}_2, \\ 1 & \text{otherwise,} \end{cases}$$

where $O(\mathbf{h}_1, \mathbf{h}_2) \in [0, 1]$ is the ratio of the area of the intersection and the area of the union of the two bounding boxes (Everingham et al., 2010). Both the loss functions not only encourage the models to predict the right category but also the right location of the object. We note that a similar experimental setup was also used by Blaschko et al. (2010).

**Dataset.** We use images of 6 different mammals (approximately 45 images per mammal) that have been previously employed for image classification (Kumar et al., 2010; Miller et al., 2012). We split the images of each category into approximately 60% for training and 40% for testing. We report results using 5 folds.

**Results.** Figure 1 shows the test loss for LSVM, ILSVM and our method using the 7 different $C$ values. The test loss is computed using the ground-truth labels and bounding boxes for the test samples. Recall that, during training, only the ground-truth labels were assumed to be known, while the bounding boxes were modeled as latent variables.

While LSVM was initially proposed for loss functions that do not depend on the value of the true latent variable, we adopted a similar approach to the CCCP algorithm for LSVM to solve the object detection problem. Briefly, we iterate over two steps: estimating the value of the latent variables and solving a convex structured SVM problem until the objective function could not be decreased below a user-specified tolerance. In our experiments, this approach provided similar results to the ILSVM method.

By incorporating the uncertainty in latent variables, our approach outperformed both LSVM and ILSVM.

Specifically, for the 0/1 loss, the best test loss (over all $C$ values) for LSVM, ILSVM and our method is $64.82 \pm 4.96$, $68.53 \pm 5.52$ and $47.76 \pm 2.53$ respectively (where the loss has been scaled to lie between 0 and 100). For the overlap loss, the best test loss is $44.93 \pm 1.84$, $47.26 \pm 3.87$ and $42.27 \pm 3.64$ respectively. While the improvement in the overlap loss is not statistically significant according to paired t-test, the improvement in the 0/1 loss is statistically significant with $p < 10^{-4}$.

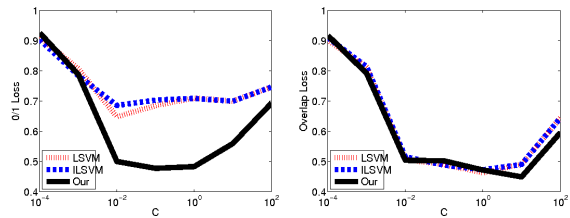

Figure 1. The average test loss over five folds (y-axis) of the object detection application for different values of C (x-axis, shown in log-scale). Left: 0/1 loss; Right: Overlap loss. Our framework outperforms both LSVM and ILSVM and provides statistically significant improvements for the 0/1 loss (see text for details).

### 6.2. Action Detection

**Problem Formulation.** The aim of this application is to learn human action models that predict the action class (for example, 'running' or 'jumping') and the location of the person present in an image. Similar to object detection, a fully supervised dataset would require annotating each training image with the person bounding box. Instead, we use image-level labels that indicate which action is being performed by a person in the image. Formally, for each sample, the input $\mathbf{x}$ is an image. The output $\mathbf{y} \in \{0, 1, \cdots, c-1\}$, where $c$ is the number of action classes. The latent variable $\mathbf{h}$ models the tight bounding box around the person in the image. The joint feature vectors are the Poselet descriptor (Maji et al., 2011) of the bounding box. We consider approximately 20 putative bounding boxes for each image, which are obtained automatically using a standard person detector (Felzenszwalb et al., 2008). The small search space for the latent variables avoids the need for approximate inference. Once again, we report results using both 0/1 loss and overlap loss.

**Dataset.** We use the PASCAL VOC 2011 'trainval' dataset (Everingham et al., 2010), which consists of approximately 2500 images of 10 different action classes. We split the images of each class into approximately 60% for training and 40% for testing, and report results using 5 folds. In addition to the detected persons, we introduce the largest ground-truth bounding box into the latent variable space.

**Results.** Figure 2 shows the test loss for the three methods, computed using ground-truth labels and bounding boxes. For 0/1 loss, the best test loss over all $C$ values for LSVM, ILSVM and our method is $93.18 \pm 1.95$, $92.89 \pm 3.70$ and $76.10 \pm 0.71$ respectively. For overlap loss, the best test loss is $70.66 \pm 0.76$, $71.33 \pm 1.14$ and $67.16 \pm 0.32$ respectively. Our method significantly outperforms both LSVM and ILSVM, as confirmed by the paired t-test with $p < 10^{-3}$.

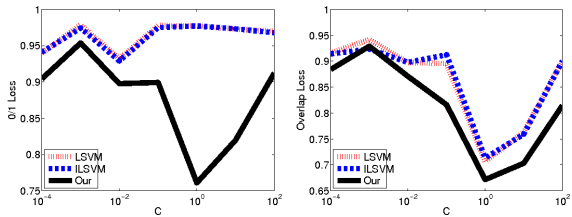

Figure 2. The average test loss over five folds (y-axis) of the action detection application for different values of $C$ (x-axis, shown in log-scale). Left: 0/1 loss; Right: Overlap loss. Our framework outperforms both LSVM and ILSVM and provides statistically significant improvements for both types of loss (see text for details).

## 7. Discussion

We proposed a novel framework for parameter estimation using weakly supervised datasets. Our framework consists of two distributions: a conditional distribution that captures the uncertainty in the latent variables, and a delta distribution that predicts the output and latent variable values. The parameters of the distributions are learned by minimizing a loss-based dissimilarity coefficient between the two distributions for all samples in the training dataset. We empirically demonstrate the benefit of our approach over previous loss-based learning frameworks using publicly available datasets of two challenging problems—object detection and action detection.

The proposed optimization requires the computation of the expected loss $\Delta_i(\mathbf{y}, \mathbf{h}|\boldsymbol{\theta})$ (shown in equation (7)) when learning the delta distribution and the loss-dependent subgradient $\mathbf{g}_t$ (shown in equation (17)) when learning the conditional distribution. In special cases (for example, low tree-width models), these terms can be computed exactly. In general, we would have to resort to one of several existing approximate inference techniques or to design customized algorithms to compute the sufficient statistics. Note that, since the conditional distribution is not used during testing, an approximate estimate of its parameters, which is able to accurately model the uncertainty in the latent variables, would suffice in practice.

**Acknowledgments.** This work is partially funded by the European Research Council under the European Community's Seventh Framework Programme (FP7/2007-2013)/ERC Grant agreement number 259112, INRIA-Stanford associate team SPLENDID, NSF under grant IIS 0917151, MURI contract N000140710747, and the Boeing company. We thank Michael Stark for proof-reading the paper, Subhransu Maji for the Poselets data, and Daniel Selsam and Andrej Karpathy for helpful discussions.